%% file: arXiv_version.tex
\documentclass{article} 
\usepackage{arxiv_version,times}
\renewcommand{\headrulewidth}{0pt}
\input{math_commands.tex}

\usepackage{hyperref}
\usepackage{url}
\usepackage{float}
\usepackage{graphicx}
\usepackage{subcaption}

\usepackage{amssymb}
\usepackage{multirow}
\usepackage[table,xcdraw]{xcolor}
\usepackage{todonotes}
\usepackage{minted}
\usepackage{verbatim}
\usepackage{svg}
\usepackage{listings}
\usepackage{xcolor}

\usepackage{enumitem}

\lstdefinelanguage{Lean}{
  morekeywords={
    def, lemma, theorem, example, inductive, structure, 
    namespace, end, open, variable, variables, 
    forall, exists, fun, match, with, if, then, else
  },
  sensitive=true,
  morecomment=[l]--,
  morecomment=[s]{/-}{-/},
  morestring=[b]"
}

\usepackage{amsmath,amsthm} 
\usepackage[most]{tcolorbox} 

\newtcbtheorem[number within=section]{theorem}{Theorem}%
{colback=blue!5!white, colframe=blue!75!black, fonttitle=\bfseries, coltitle=black,
  enhanced, attach boxed title to top left={yshift=-2mm,xshift=2mm},
  boxed title style={colback=blue!50!white}}{th}

\newenvironment{proofbox}{%
  \begin{tcolorbox}[colback=white, colframe=green!60!black, title=Proof]}%
  {\end{tcolorbox}}

\newenvironment{theorembox}{%
  \begin{tcolorbox}[colback=white, colframe=blue!60!black, title=Theorem]}%
  {\end{tcolorbox}}

\title{ProofFlow: A Dependency Graph Approach to Faithful Proof Autoformalization}

\author{Rafael Cabral$^1$, Tuan Manh Do$^1$, Yu Xuejun$^1$,  Wai Ming Tai$^1$, Zijin Feng$^2$, Xin Shen$^1$  \\
$^1$Huawei Celia Team \\
$^2$Huawei Noah’s Ark Lab \\ 
Corresponding email: shenxin19@huawei.com}

\definecolor{forestgreen}{RGB}{34,139,34}

\iclrfinalcopy 
\begin{document}

\maketitle
\begin{abstract}
Proof autoformalization, the task of translating natural language theorems and proofs into machine-verifiable code, is a critical step for integrating large language models into rigorous mathematical workflows. Current approaches focus on producing executable code, but they frequently fail to preserve the semantic meaning and logical structure of the original human-written argument. To address this, we introduce \textsc{ProofFlow}, a novel pipeline that treats structural fidelity as a primary objective. \textsc{ProofFlow} first constructs a directed acyclic graph (DAG) to map the logical dependencies between proof steps. Then, it employs a novel lemma-based approach to systematically formalize each step as an intermediate lemma, preserving the logical structure of the original argument. To facilitate evaluation, we present a new benchmark of 184 undergraduate-level problems, manually annotated with step-by-step solutions and logical dependency graphs, and introduce \textsc{ProofScore}, a new composite metric to evaluate syntactic correctness, semantic faithfulness, and structural fidelity. Experimental results show our pipeline sets a new state-of-the-art for autoformalization, achieving a \textsc{ProofScore} of 0.545, substantially exceeding baselines like full-proof formalization (0.123), which processes the entire proof at once, and step-proof formalization (0.072), which handles each step independently.
Our pipeline, benchmark, and score metric are open-sourced to encourage further progress at \githublink.




\end{abstract}


\section{Introduction}\label{sect:introduction}

\input{010intro}

\section{Background and related work}\label{sect:background}

\input{020relatedwork}

\section{Proof autoformalization}\label{sect:autoformalization}

\input{030autoformalization}

\section{Scoring Proof autoformalization and error detection} \label{sect:score}

As discussed in Section \ref{sect:introduction}, a faithful autoformalization must satisfy three key properties: (1) Structural Fidelity, which ensures the proof's dependency graph is preserved; (2) Syntactic Correctness, which ensures the output is verifiable code without compilation errors; and (3) Semantic Faithfulness, which ensures each formalized statement accurately preserves the precise mathematical meaning of its original natural language statement.



\subsection{\proofscore}

To evaluate the effectiveness of our  \pipeline \ pipeline, we introduce \proofscore, a single unified score that synthesizes these three criteria.

\textbf{Structural Fidelity} is evaluated by checking whether the dependencies for a given node are valid. For a node $v_i$, we check if its estimated dependencies, $D_{\text{est}}(v_i) = \{u \in V_{\text{est}} \mid (u, v_i) \in E_{\text{est}}\}$, match the dependencies in a ground truth graph. We permit several valid graphs ($\mathcal{G}_{\text{true}}$) because the level of granularity can vary. For example, the calculation $1 + 13 + 5 = 14 + 5 = 19$ could be broken down into either one or two steps, depending on the user's granularity preference.


\textbf{Syntactic Correctness} is denoted by $c_i \in \{0, 1\}$, where $c_i=1$ if the formalization of node $v_i$ is free of Lean 4 compilation errors at the end of the Tactic Completer step and $c_i=0$ otherwise.

\textbf{Semantic Faithfulness} is assessed by adapting the ``LeanScore'' metric from~\cite{xuejun2025mathesis}, which was originally designed to evaluate the semantic faithfulness of a theorem statement formalization. A detailed description is given in Appendix \ref{sect:leanscore}. This metric provides a faithfulness score, $f_i \in [0, 1]$, for each node $v_i$ of the dependency graph, and this score measures the semantic equivalence between the input NL statement and its corresponding formalized Lean 4 lemma. The higher the score, the more faithful the Lean 4 lemma devotes to the input NL statement. 



The final unified \proofscore, for a proof with $n$ steps is computed as:
$$\text{\proofscore} = \frac{1}{n}\sum_i^n f_i \ c_i  \ I[D_{\text{est}}(v_i) \in \mathcal{D}_{\text{true}}(v_i) ],$$ 
where $\text{ProofScore} \in [0,1]$, and $\mathcal{D}_{\text{true}}(v_i)$ is the set of valid dependencies for node step $v_i$. When the GraphBuilder is configured to enforce the DAG structure, structural fidelity is guaranteed for all nodes. Otherwise, we use the a high-performance proprietary LLM with a specialized prompt to verify structural fidelity. This same LLM model is also used to check for semantic faithfulness, utilizing prompts available within the \pipeline~package. 

\subsection{Error analysis}\label{sect:error_analysis_main}

\begin{figure}[t]
\vspace{-0.1cm}
    \centering
    \includegraphics[width=0.8\linewidth]{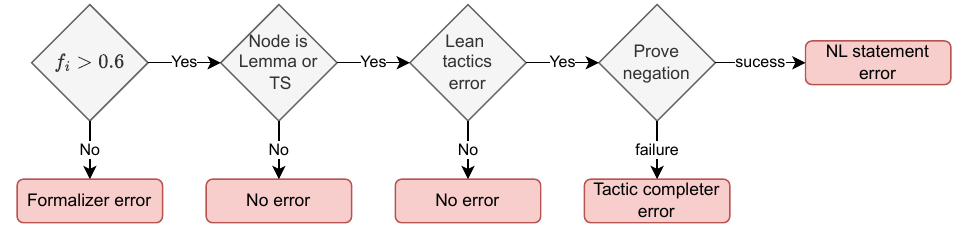}
    \caption{Flowchart illustrating the error detection mechanism. This is applied lemma by lemma and $f_i$ is the semantic faithfulness of node $v_i$ (between 0 and 1).}
    \label{fig:error_decision}
    \vspace{-1.25\baselineskip}               
\end{figure}

\pipeline \ also has an error detection system operating at each proof step. This process (Figure \ref{fig:error_decision}) determines the error source, which can be the formalizer, tactic completer, or original natural language proof. First, a proof node's semantic faithfulness score is compared against a 0.6  threshold, following~\cite{xuejun2025mathesis}. If it passes, the system checks if it's a provable statement (lemma or theorem solution). Otherwise, the process for that node ends. For provable statements, the tactic completer attempts to apply Lean 4 tactics. Success means the proof step is complete with no errors. If the tactics fail, the LLM tries to prove the negation of the statement.  If the negation can be proven, then the original natural language statement is deemed imprecise. Otherwise, the error is considered to be with the tactic completer LLM, which is incapable of either proving or disproving the step.

\vspace{-0.125cm}
\section{\benchmark} \label{sect:benchmark}
\vspace{-0.125cm}

\input{050benchmark}

\vspace{-0.125cm}
\section{Comparison study} \label{sect:comparison}
\vspace{-0.125cm}
To evaluate our \pipeline \ pipeline, we conduct a  comparison study with three main objectives. First, we assess the effectiveness of formalizing proofs using our high-level, lemma-based approach (\pipeline) versus low-level tactics. This compares the ``Lemma Formalization" (ours) with the ``Tactic Formalization" (existing), as depicted in Figure~\ref{fig:autoformalization_task}. 
Secondly, to examine the role of explicit dependency management in ensuring structural fidelity to the original proof, we compare the DAG-enforcing version of \pipeline \ with a variant where this is relaxed (noDAG). Finally, we conduct the error analysis of Section \ref{sect:error_analysis_main}, to detect the source of errors in the \pipeline \ pipeline variants. 

\subsection{Experimental Settings}
\label{subsec:exp_settings}

\textbf{\pipeline}: We consider the \pipeline \ variant that explicitly enforces the correct dependency graph as our main method, referred to as \pipeline \ DAG. To evaluate the role of this dependency enforcement, as an ablation study, we also consider a \pipeline \ noDAG version, where such mechanism is relaxed. That is, in \pipeline \ noDAG, all previous lemmas and premises are provided when formalizing each step, similarly to how dependency is handled by most prior work.

\textbf{Existing methods}: We compare our pipeline against existing tactic-based formalization methods: (1) ``\fullproof" autoformalization \citep{lu2024process}, which calls the LLM once to formalize the entire theorem and proof using Lean 4 tactics, and (2) ``\steplevel" autoformalization \citep{hu2025stepproofstepbystepverificationnatural}, which formalizes one proof step at a time into Lean 4 tactics.

\textbf{Thinking modes}: All methods were evaluated in both thinking and non-thinking modes to provide a more comprehensive assessment. For existing methods, a high-performance proprietary LLM with thinking mode turned on and a fast-response proprietary LLM with thinking mode turned off are used for the thinking and non-thinking modes\footnote{The specific commercial models used in our experiments are anonymized. We describe them by their general performance characteristics to provide context while adhering to publication guidelines.}. For our \pipeline \ pipeline, the Graph Builder always uses a high-performance proprietary LLM with thinking mode turned on;  the thinking mode utilizes Goedel-Formalizer-2 32B and Goedel-Prover-V2 32B \citep{lin2025goedelproverv2scalingformaltheorem} as the Formalizer and the Tactic Completer, while non-thinking mode utilizes a fast-response proprietary LLM with thinking mode turned off and DeepSeek-Prover-V2 671B \citep{ren2025deepseekproverv2advancingformalmathematical} for the two components. All LLM models were given appropriate system prompts.

\textbf{Evaluation metrics:} Our main evaluation metric in \proofscore, supplemented by measures of syntactic correctness, both at step-level and proof-level. Step-level syntactic correctness of the generated Lean 4 code is measured at two key stages: after the Formalizer step (Formalizer accuracy) and after the Tactic Completer step (Tactic accuracy). We also report proof-level ``correct syntax'', whether all of the formalizer and solver steps are syntactically valid.  All syntactic correctness checks are verified by LeanServer \citep{santos2025kiminaleanservertechnical} with Lean version v4.15.0. We also adapted \proofscore \ (Sect.~\ref{sect:score}), so it could evaluate the \fullproof \ and \steplevel \  pipelines (see Appendix~\ref{app:leanscore_competitors} for details). Finally, we report the elapsed time and the number of generated tokens. Since our experiments use both local and API-based model deployments, inference speeds vary. Therefore, the number of generated tokens is a more meaningful indicator of the effort required.

\textbf{Data and reproducibility}: We utilized the problems in the \benchmark \ benchmark for this comparison study. Reproducible code is provided as part of the \pipeline \ package.

\subsection{Empirical Results}
\label{subsec:results}

The results evaluated at a Pass@5 rate are presented in Table~\ref{table:comparison_table_1}. This Pass@5 setting allows for up to five self-correction retries for each stage in the pipelines. If a generated step produces syntactically incorrect code, the model is re-prompted with the error and its previous output, giving it an opportunity to fix the mistake. The results for Pass@1 and 3 are show in Appendix \ref{app:comparison_results}.

\input{tab_comparison_main}

\textbf{Existing Methods Comparison:} The results clearly demonstrate that our pipeline outperforms existing methods (\fullproof \ and \steplevel), achieving the highest proof-level syntax passing rate (0.375) and \proofscore \ (0.545). The \fullproof \ method calls the LLM once for the entire proof formalization and therefore we only report proof-level results. This approach achieves low syntax passing rates (0.027 for thinking and 0.141 for non thinking modes), with the most frequent errors being the misuse of unknown theorem names and tactic failures. The \steplevel \ method exhibits the lowest proof-level syntax passing rates (0.005) because the LLM struggles to maintain correct and consistent indentation in Lean 4 across steps. Also, once an error appears in one step, subsequent steps cannot achieve syntactic correctness. The poor syntax passing rates of both \fullproof \ and \steplevel \ approaches result in correspondingly low \proofscore \ values, given that this metric incorporates syntactic correctness as a component.

\textbf{Ablation Study}: The \proofscore \ evaluation metric establishes the DAG version of \pipeline \ as the top-performing method for proof autoformalization. This is most evident in the ``thinking" mode, where the DAG variant achieves a \proofscore \ of 0.545, compared to the noDAG's 0.417, due to improved structural fidelity and syntactic correctness. While the DAG also outperforms the noDAG variant for all syntax metrics in the thinking mode, we observed a higher syntax passing rate for the noDAG in the non thinking setting. This is because the noDAG variant provides all previous steps as known conditions to the LLM, and this extra information can lead to higher syntactic passing rates for weaker models, but often results in a logically inconsistent formal proof structure, as shown by the lower \proofscore \ averages. This finding underscores the critical importance of maintaining structural fidelity.

The error analysis, shown in Table~\ref{tab:error_analysis_main}, identifies the Formalizer as the primary source of failure, accounting for 32\% to 47\% of all autoformalization outcomes, depending on the pipeline configuration. These failures are predominantly semantic, stemming from discrepancies between the natural language and the formalized Lean 4 code, despite the high syntactic correctness (Table~\ref{table:comparison_table_1}). Our most robust configuration, which uses the \pipeline~DAG architecture in thinking mode, has 53.3\% error-free proof steps and clearly outperforms the 42.8\% rate of the baseline (noDAG) version. This confirms that while the DAG architecture improves performance, improving semantic preservation during the formalization stage remains a challenge for future work.

The \pipeline's error detection process goes beyond simple checks, being able, in some instances, to identify a spectrum of flaws in the initial natural language proof. Examples are provided in Appendix~\ref{sect:nl_errors}. These detected issues range from blatant algebraic errors to subtle ambiguities and unstated assumptions. By flagging these mistakes, pipelines like \pipeline \ can provide feedback that not only helps correct errors but also strengthens the overall rigor and clarity of the proof.


\input{tab_error_analysis}

\section{Discussion}

Our results demonstrate the importance of preserving a proof's structural fidelity. The \pipeline \ lemma-based approach significantly outperforms both monolithic (``\fullproof'') and sequential (``\steplevel'') methods, which falter by tackling excessive complexity at once or by imposing a linear structure unfaithful to the proof logic. Our method succeeds by using a lemma-based structure that explicitly models the proof's dependency graph. This approach deconstructs the problem into manageable, logically-constrained steps. It guides the LLM along the author's intended path, preventing logical ``shortcuts'' that undermine other approaches. However, performance hinges on the ``Formalizer'' step, where poor semantic preservation remains the primary bottleneck. 

Ultimately, our work demonstrates that proof autoformalization is an achievable goal: we achieved 37.5\% proof-level syntactic accuracy on undergraduate problems, a significant leap from the previous benchmark of 6.10\% on simpler middle-school mathematics \citep{hu2025stepproofstepbystepverificationnatural}. More importantly, we establish that for these tools to be truly useful to mathematicians, they must faithfully represent the structure and logic of the human-created arguments they seek to formalize. Addressing the current bottlenecks could lead to practical tools that function as a proof ``auto-correct'' for mathematicians, pinpointing errors and ambiguities directly in natural language and in real time.

\newpage
\bibliographystyle{iclr2026_conference}
\bibliography{arXiv_version}

\newpage
\appendix
\section{Appendix} \label{sect:appendix}
\input{099appendix}

\end{document}

%% file: math_commands.tex
\usepackage{amsmath,amsfonts,bm}

\newcommand{\pipeline}{\textsc{ProofFlow}}

\newcommand{\proofscore}{\textsc{ProofScore}}
\newcommand{\fullproof}{\textsc{Full Proof}}
\newcommand{\steplevel}{\textsc{Step Proof}}
\newcommand{\benchmark}{\textsc{ProofFlowBench}}
\newcommand{\githublink}{\url{https://github.com/Huawei-AI4Math/ProofFlow}}

\usepackage{xcolor}

\def\eqref#1{equation~\ref{#1}}

\def\1{\bm{1}}

\DeclareMathAlphabet{\mathsfit}{\encodingdefault}{\sfdefault}{m}{sl}
\SetMathAlphabet{\mathsfit}{bold}{\encodingdefault}{\sfdefault}{bx}{n}



%% file: 010intro.tex
The effort to automate mathematical reasoning is advancing on two key fronts. Recent advances in large language models (LLMs) have greatly enhanced their ability to solve mathematical problems~\citep{liang2025towards}. Meanwhile, symbolic engines such as Lean \citep{moura2021lean} and Isabelle~\citep{hales2017formal} provide machine-verifiable frameworks that enforce strict logical correctness. LLM-based automated theorem provers (ATPs), such as Goedel-Prover~\citep{lin2025goedelproverv2scalingformaltheorem} and Kimina-Prover~\citep{kimina_prover_2025}, generate formal proofs for problems that are written in a symbolic language, which are then checked by the symbolic engine to ensure logical correctness.


This paper studies automated proof formalization: the task of faithfully translating the natural language theorem and proof into a machine-verifiable formal representation. This task is distinct from the aforementioned ATPs, where the goal is to discover a proof from an already formalized problem. Here, the goal is to translate an existing proof, a crucial step for verification in real-world mathematical workflows.  After composing a proof, a mathematician may wish to formalize it to verify its correctness, uncover missing assumptions, or fill logical gaps. Given the steep learning curve of formal languages like Lean, a system that can automate this translation is highly desirable. The manual effort these automated proof formalizers seeks to automate is immense, as demonstrated by landmark efforts such as the Flyspeck project, which took over 20 years to formalize the Kepler conjecture~\citep{hales2017formal}; the Blue-Diamond project for the Polynomial Freiman-Ruzsa conjecture~\citep{tao2023pfr}; and the ongoing formalization of Fermat's Last Theorem~\citep{buzzard2025lean}, among other projects \citep{strongpnt_github,scholze2021liquid,gonthier2013machine}. 


\begin{figure}[ht!]
\vspace{-0.4cm}
\centering
\includegraphics[width=1\linewidth]{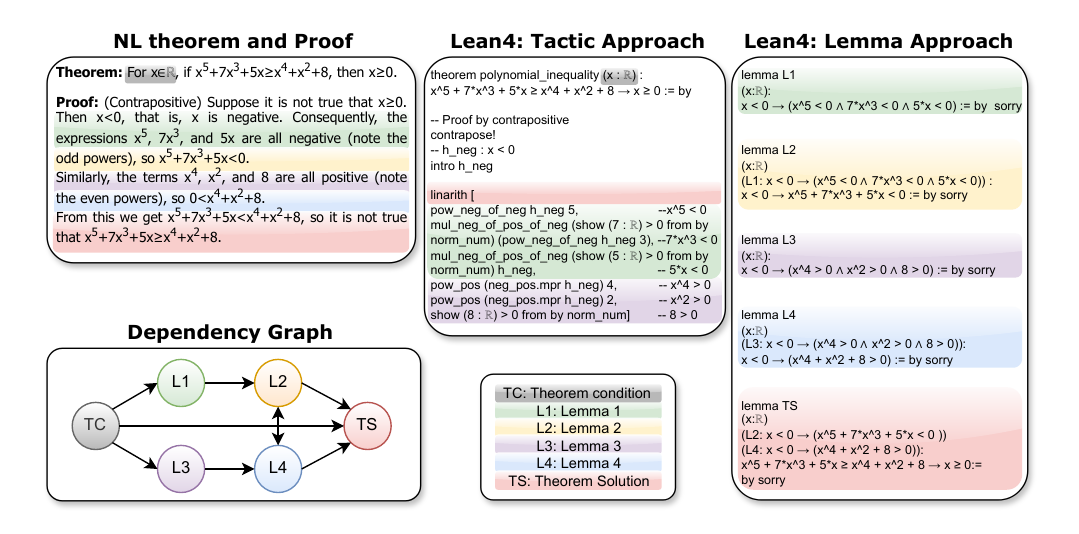}
\caption{Comparison of our \pipeline \ ``Lemma Approach" \ and the common ``Tactic Approach" in formalizing a natural language (NL) theorem and its proof into Lean 4 code. The Lemma Approach directly mirrors the sequence of steps and their dependencies in the NL proof. By contrast, the Tactic Approach produces tactics that fail to adhere to the structure of the initial NL proof.}
    \label{fig:autoformalization_task}
\vspace{-1.25\baselineskip}               
\end{figure}


A key challenge in autoformalization is the fundamental mismatch between the flexible and high-level nature of human language and the rigid, low-level syntax of formal systems. To bridge this gap, ATPs typically generate ``tactics'' for formal proof assistants like Lean 4. However, this tactic-based approach often fails to preserve the logical structure of the original human argument. For instance, as shown in Figure~\ref{fig:autoformalization_task}, under ``Lean 4: Tactic Approach", the generated tactics do not follow the sequence of steps in the natural language proof. Furthermore, a single tactic, \textit{linarith}, consolidates three distinct steps from the original proof (TS, L2, and L4). This disparity creates two major issues. First, not all natural language mathematical expressions can be directly translated into low-level tactics, which can cause the formalization process to fail. Second, even if an automated system generates a verifiable proof, it may take shortcuts or skip intermediate steps, arriving at the correct conclusion without mirroring the explicit, step-by-step reasoning of the original proof. This makes it difficult to verify that the formal proof truly captures the human's intended logic.

To address these issues and ensure faithful autoformalization, we propose a novel approach that avoids direct translation into limited, low-level formal tactics. Instead, we deconstruct the natural language proof into a sequence of structured, high-level lemmas, as illustrated by the ``Lemma Approach'' in Fig.~\ref{fig:autoformalization_task}. The key advantage of this method is that it enables us to explicitly follow the sequence of steps in the original proof with low friction\footnote{While we currently use the placeholder ``by sorry" for unproven parts, our ultimate goal is to generate Lean tactics to prove each lemma and the final solution.}. This stands in sharp contrast to the conventional tactic-based approach, which is often convoluted and hinders mirroring the human proof structure.
Second, our lemma-based approach preserves the logical structure of the natural language proof by explicitly defining dependencies. As shown in Figure~\ref{fig:autoformalization_task}, the final step, lemma TS (the theorem's solution), depends only on lemma \textsc{L2} and lemma \textsc{L4}. This explicit dependency management is crucial in enforcing structurally faithful formalization, which is often not achieved in a standard tactic-based workflow. For instance, a common tactic like ``have" in Lean 4 feeds the entire preceding context to each new step, not just the specific premises required by the original proof logic. In contrast, our approach of explicitly defining how lemmas depend on one another prevents the system from incorrectly using unintended dependencies, a common failure in tactic-based proving. These two innovations, high-level lemma generation and explicit dependency tracking, form the core of our new autoformalization pipeline, \pipeline.


Another primary challenge is defining what constitutes a ``faithful'' formalization. In existing work, researchers either focus on syntactic correctness \citep{hu2025stepproofstepbystepverificationnatural}, i.e., no compilation errors, or use a simple BLEU score for semantic measurement \citep{poiroux2024improving,wu2022autoformalization}, while ignoring the structural fidelity of the proof. To properly evaluate proof formalizations, we propose viewing a proof not as a monolithic block of text, but as a structured sequence of theorem conditions, definitions, and lemmas that form a logical progression toward the final theorem solution or solutions (see Figure \ref{fig:autoformalization_task}). Based on this, we introduce a new and more comprehensive proof autoformalization scoring metric (\proofscore), and we also address the lack of an advanced benchmark by providing a new university-level dataset tailored for this task (\benchmark).

Drawing on our new framework, we make the following key contributions:
\begin{itemize}[leftmargin=*, itemsep=0.25em]
    \item \pipeline{:} In Section \ref{sect:autoformalization}, we propose a novel pipeline for translating natural language theorems and proofs into structured and formal Lean code, ensuring the preservation of the proof's logical structure. When a formalization step fails, the pipeline identifies the error source, be it in the formalization, the tactic completion process, or the initial NL statement, thereby alerting mathematicians of potential flaws in their original proof.
    \item \proofscore: Section \ref{sect:score} introduces a new and comprehensive scoring method to evaluate the quality of autoformalized proofs. This metric is the first to explicitly measure syntactic correctness, semantic faithfulness, and structural fidelity, providing a more complete assessment than existing methods.
    \item \benchmark: In Section \ref{sect:benchmark}, we present a new, manually curated benchmark dataset for proof autoformalization, containing a collection of 184 undergraduate level problems.
    \item Comparative Study: Section \ref{sect:comparison} presents an empirical study using state-of-the-art models to evaluate our structure-aware pipeline, \pipeline, against alternative strategies. The results show that \pipeline \ has significantly higher proof autoformalization quality.
\end{itemize}

%% file: 020relatedwork.tex
\textbf{Proof assistants and automatic theorem proving:} Proof assistants like Isabelle \citep{Paulson1994Isabelle}, Lean 4 \citep{moura2021lean}, and Coq \citep{barras1997coq} are software environments for developing and verifying mathematical proofs.  Proofs constructed within these systems are what we call ``formal" proofs, distinguishing them from informal or natural language proofs written in standard mathematics (e.g., in \LaTeX).  The user's workflow involves interactively applying tactics, which are small programs that perform logical inferences, to solve the theorem's goals \citep{jiang2023draft}. Despite their power, a steep learning curve and the significant manual effort required by their rigid syntaxes have hindered widespread adoption \citep{zhou2024donttrustverify}. Recently, Large Language Models (LLMs) have emerged as powerful automated theorem provers (ATPs) capable of generating complete formal proofs from already formalized theorem statements \citep{shang2025stepfun, lin2025goedelproverv2scalingformaltheorem, wang2025kimina, ren2025deepseek}. Proof Agents have further advanced this ability \citep{chen2025seed, zhou2025solving, baba2025prover}. In the context of Figure \ref{fig:autoformalization_task}, the task of an ATP is to automatically replace the placeholder command ``sorry" with formal tactics to complete the proof. 

\textbf{Autoformalization with LLMs:} For a fully automatic theorem-proving system, mathematical problems originating in natural language must first be translated into a formal language, a process known as autoformalization.  Historically, autoformalization efforts have primarily focused on translating theorem statements, and not the natural language proof \citep{huang2025formarl, wu2025stepfun, liu2025atlas,yu2025formalmath, poiroux2024improving, pathak2024gflean,patel2023new}, often to support the training of automated provers \citep{lin2025goedelproverv2scalingformaltheorem, wang2025kimina}. 
A different approach uses informal proof sketches to guide an LLM's search for a formal proof \citep{cao2025reviving, zhou2024donttrustverify, jiang2023draftsketchproveguiding}. In these methods, the natural language proof sketch is not the target of formalization itself. Rather, it serves as a high-level guide, often interleaved as comments within the formal code to steer the generation process. This guidance technique is also employed by recent ATPs like DeepseekProver-V2 \citep{ren2025deepseek} and Goedel Prover V2 \citep{lin2025goedelproverv2scalingformaltheorem}.

\textbf{Proof Autoformalization:}
In the literature of proof autoformalization, most existing attempts directly translate entire proofs using LLMs \citep{gao2024herald, lu2024process, cunningham2023towards}. This approach, however, remains highly challenging due to frequent syntactic errors and it often produces outputs that are not semantically trustworthy.  Step-level formalization, which involves solving the autoformalization step-by-step based on the proof's logical steps, has been explored by \cite{hu2025stepproofstepbystepverificationnatural} in Isabelle. However, this approach suffers from two limitations. First, they treated all previous proof steps as valid premises for the current step, a significant simplification that overlooks the proper logical structure. Second, their evaluations focused solely on syntactic correctness while overlooking semantic consistency. 



%% file: 030autoformalization.tex

\begin{figure}[t]
    \centering
    \begin{subfigure}[t]{0.54\linewidth}
        \centering
        \includegraphics[width=\linewidth]{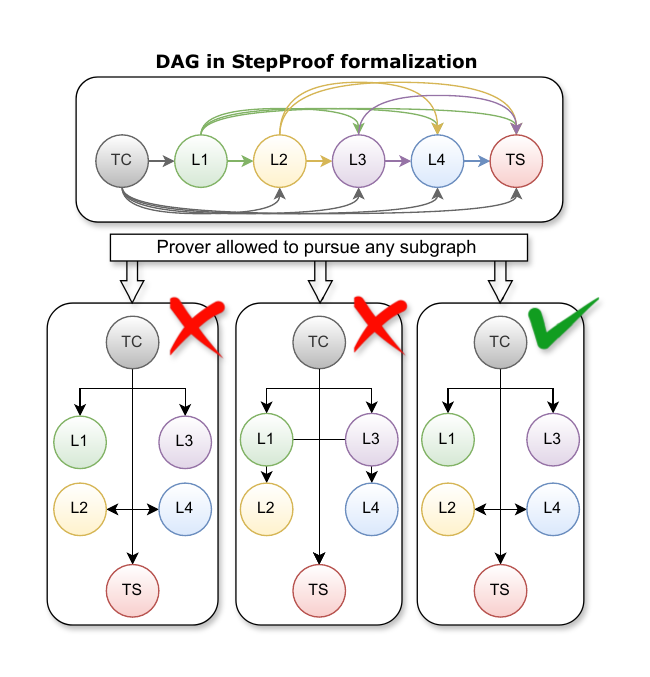}
        \caption{} 
        \label{fig:DAG_issue}
    \end{subfigure}
    \hfill
    \begin{subfigure}[t]{0.45\linewidth}
        \centering
        \includegraphics[width=1\linewidth]
        {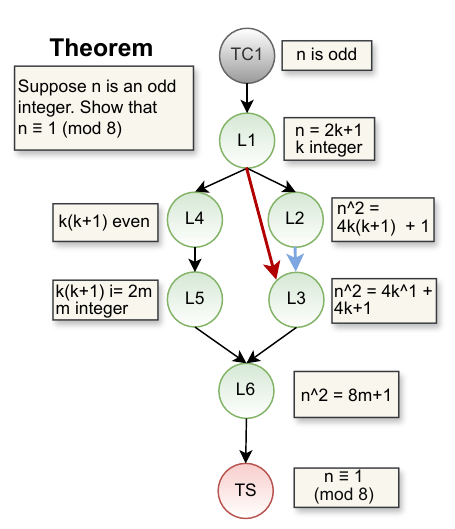}
        \caption{} 
        \label{fig:cheating}
    \end{subfigure}
    \caption{Comparing structural fidelity in automated proof generation. (a) A scenario for the problem in Figure \ref{fig:autoformalization_task}, where the dependency graph fails to adhere to the structure of the original proof. (b) This benchmark problem was intentionally formalized without enforcing a DAG. This approach sacrificed structural fidelity, by reusing lemma \textsc{L1} to prove \textsc{L3}, thereby rendering \textsc{L2} redundant.}
    \label{fig:side_by_side}
\end{figure}

Although most prior work formalizes entire proofs at once, a notable exception is the ``\steplevel" method \citep{hu2025stepproofstepbystepverificationnatural}. This approach assumes each proof step depends on all preceding steps, a simplification that can lead to unintended consequences. This unfaithful dependency structure may cause an ATP to take a ``shortcut," using only the initial theorem conditions or an incorrect subset of lemmas to construct a valid proof that does not follow the logic of the original NL proof, as depicted in Figure \ref{fig:DAG_issue}. A clear example found in \benchmark \ is illustrated in Figure~\ref{fig:cheating}. In the natural language proof, step L3 directly uses the outcome of L2, which is faithfully followed by our DAG-enforcing pipeline. By contrast, an ablated version of our pipeline which lacks the mechanism to enforce the correct dependencies (noDAG) reuses the outcome of L1 to prove L3, despite having proven L2. Thus, it not only dissipates step L2 but also disregards the structure of the original proof. More details and examples are found in Appendix~\ref{app:examples}.

 

If the sole objective is to find any correct formal proof for a theorem, which is often the sole goal of ATPs, these behaviors are not problematic. However, for proof autoformalization, they represent a critical failure of faithfulness. Our goal is to ensure that each step is proven only taking into account the correct set of previous lemmas and theorem conditions specified by the original proof's logic. Enforcing this correct dependency structure offers significant practical advantages as well. It improves efficiency by constraining the autoformalizer's search space, which can lead to fewer spent tokens and faster verification. 


\vspace{-2mm}

\subsection{New workflow for proof autoformalization} \label{sect:pipeline}

\begin{figure}[ht!]
    \centering
    \includegraphics[width=1\linewidth]{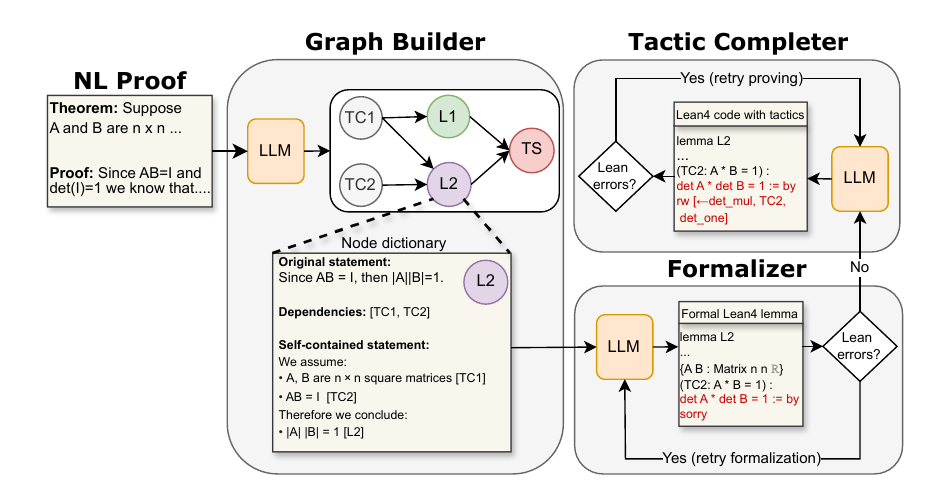}
    \caption{Our proof autoformalization pipeline with three parts: (1) Graph builder; (2) Lemma Formalizer; and (3) Tactic Completer. Lean errors are verified by the Lean 4 compiler.}
    \label{fig:pipeline}
    \vspace{-1.25\baselineskip}               
\end{figure}

\vspace{-2mm}

As shown in Figure \ref{fig:pipeline}, our method enforces a correct dependency graph through a three-stage pipeline, leveraging LLMs at each step to bridge the gap between natural language (NL) and formal proof code. The first stage, Graph Builder, constructs a dependency directed acyclic graph (DAG) from the original proof. Next, the Formalizer uses an LLM to translate each proof step into formal Lean code. Finally, the Tactic Completer fills in the necessary tactics to complete the Lean proof. The specific LLM models used at each stage are detailed in Section \ref{sect:comparison}.

\textbf{1. Graph Builder:}
This step parses the natural language (NL) theorem and its proof to construct a dependency graph with LLM. Formally, this graph is a DAG, $G = (V, E)$, where $V$ is the set of all nodes and $E \subseteq V \times V$ is the set of directed edges. The nodes are partitioned into disjoint sets: $V = V_{TC} \cup V_D \cup V_L \cup V_{TS}$, representing Theorem Conditions, Definitions, Lemmas, and Theorem Solutions, respectively. Each edge $(u, v) \in E$ signifies that node $u$ is a prerequisite for proving the statement of node $v$. From the theorem statement, we extract nodes for theorem conditions and theorem solutions.  From the proof statement, we extract nodes for lemmas and extra definitions. Each node is assigned its original NL statement, its dependencies, and a self-contained NL statement that provides a complete description of the current proof step (see Figure \ref{fig:pipeline}). To ensure the graph's validity, the system checks for forward references and cycles and verifies that every node, except the theorem solutions, has an outgoing edge. If the check fails, we task the LLM to improve the graph.


\textbf{2. Formalizer:}
\label{para:formalizer}
For each node in the graph, the LLM formalizes its self-contained NL statement into Lean 4 code. This process is iterative: generated errors are fed back to the LLM for correction. At this stage, each lemma is finalized with the ``by sorry" placeholder, as  tactics are not yet applied.

\textbf{3. Tactic Completer:}
The final step completes the proofs for the lemmas by replacing the ``by sorry" placeholders in the formalized Lean 4 code with the appropriate Lean 4 tactics.





To streamline the entire workflow illustrated in Figure \ref{fig:pipeline}, we developed a user-friendly Python package for \pipeline. This package automates the complete process, allowing users to provide an informal theorem and proof with just a few lines of code. The package then autonomously executes the full workflow. It also integrates \proofscore \ evaluation and our \benchmark \ benchmark for comprehensive assessment. The package is available \footnote{A public GitHub repository containing the package will be made available upon paper acceptance.} at \githublink.

\begin{figure}[ht!]
    \centering
    \includegraphics[width=1\linewidth]{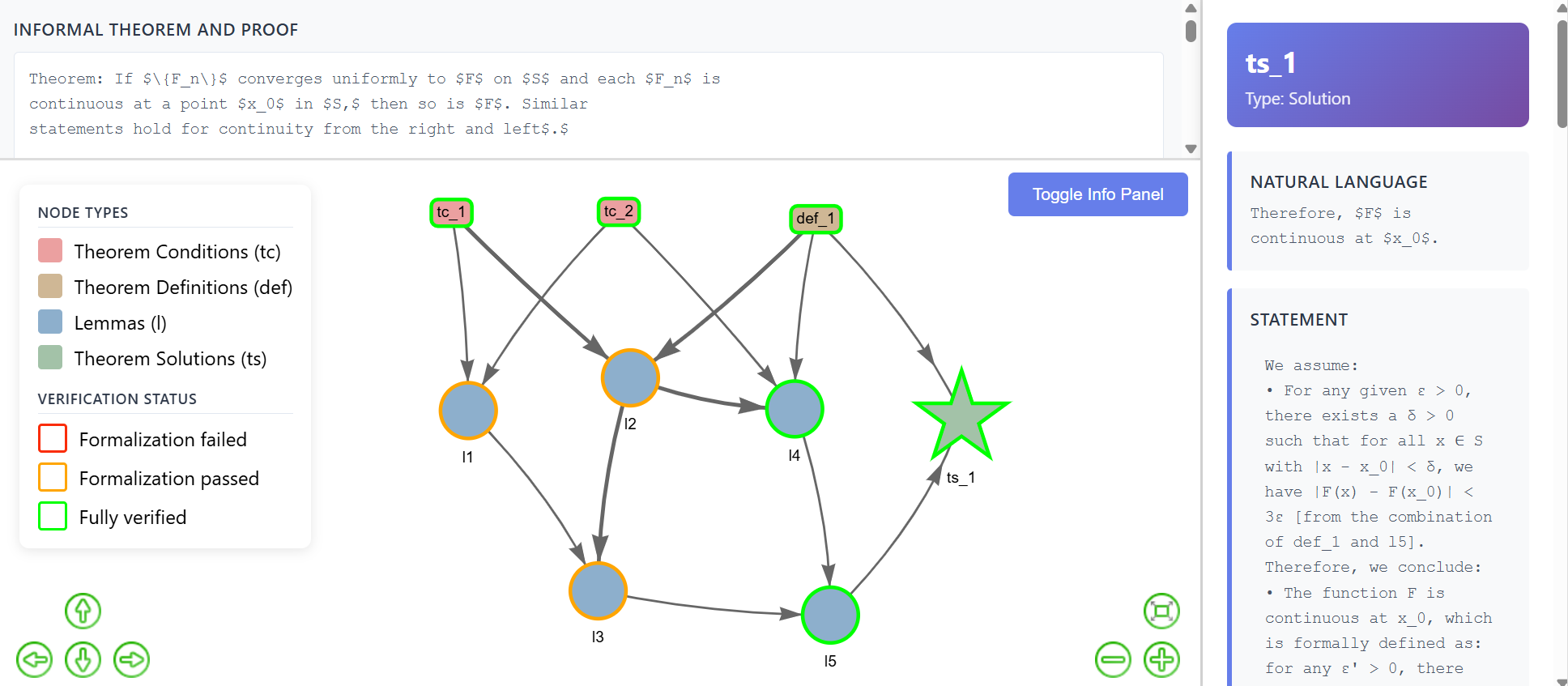}
    \caption{An example of the interactive visualization generated by \pipeline. Node contours signify the outcome of each step: Red for a formalization error, orange for formalized statement without Lean 4 tactics, and green for formalized statement with Lean 4 tactics. \vspace{-0.35cm}}
    \label{fig:demo}
    
\end{figure}

The final output of our package is an interactive diagram that visualizes the proof graph, as seen in Figure \ref{fig:demo}. Users can click on each node to get detailed information about its results. The contours of each node provide an at-a-glance summary of the outcomes, indicating whether its formalization and solving steps were successful. This allows users to immediately assess the progress of the autoformalization and pinpoint the locations in the proof graph where manual effort is needed.

%% file: 050benchmark.tex
Existing mathematical benchmarks are often limited to pure calculation and are designed to test problem solving accuracy. Previous natural language proof datasets are also not self-contained, often referencing external sources for non-standard theorems \citep{welleck2021naturalproofsmathematicaltheoremproving}. Furthermore, they focus on specific topics and sometimes also including calculation problems instead of proofs \citep{sheng2025solving}. To address these limitations, we introduce a new benchmark, \benchmark, to specifically evaluate automated proof formalization pipelines. The benchmark, provided as part of the \pipeline \ package,  consists of 184 undergraduate-level mathematics theorems and proofs in natural language from six key areas: number theory and algebra (27), real analysis (42), inequality (36), probability and set theory (31), complex analysis (25), and sequences and series (23). To construct this dataset, we adapted 63 problems from the NaturalProofs benchmark \citep{welleck2021naturalproofsmathematicaltheoremproving} and 36 problems from the IneqMath benchmark \citep{sheng2025solving}. 

The benchmark also contains the natural language proofs divided into proof steps, and the respective dependency graphs (DAGs), which can be used to evaluate structural fidelity. On average, each problem's graph consists of 8.4 total nodes (more statistics provided in Appendix \ref{sect:benchmark_stats}).  The proof graphs have been manually validated. However, we emphasize that for the same proof different proof graphs are possible depending on the desired level of detail of each proof node.





%% file: tab_comparison_main.tex
\begin{table}[t]
\vspace{-0.2cm}
\caption{Performance metrics under the Pass@5 setting on our 184-problem benchmark. Step-level averages are computed over all individual steps, while proof-level averages are computed per proof. Entries marked with ``/" indicate not applicable.\vspace{-0.35cm}}
\label{table:comparison_table_1}
\begin{center}
\renewcommand{\arraystretch}{1.25}
\begin{tabular}{cc|cc|cccc}
\multicolumn{1}{c}{} & \multicolumn{1}{c|}{} & \multicolumn{2}{c|}{\bf Step-Level} & \multicolumn{4}{c}{\bf Proof-Level} \\
\multicolumn{1}{c}{\bf Pipeline} & \multicolumn{1}{c|}{\bf \begin{tabular}[c]{@{}c@{}}Think \\ mode\end{tabular}} & \multicolumn{1}{c}{\bf \begin{tabular}[c]{@{}c@{}}Form. \\ accuracy\end{tabular}} & \multicolumn{1}{c|}{\bf \begin{tabular}[c]{@{}c@{}}Tactic \\ accuracy\end{tabular}} & \multicolumn{1}{c}{\bf \begin{tabular}[c]{@{}c@{}}Proof \\ Score\end{tabular}} & \multicolumn{1}{c}{\bf \begin{tabular}[c]{@{}c@{}}Correct \\ syntax\end{tabular}} & \multicolumn{1}{c}{\bf \begin{tabular}[c]{@{}c@{}}Time \\ (mins)\end{tabular}} & \multicolumn{1}{c}{\bf \begin{tabular}[c]{@{}c@{}}Output \\ tokens (k)\end{tabular}} \\ 
\hline \\
\multirow{2}{*}{\bf \begin{tabular}[c]{@{}c@{}}\pipeline \\ DAG\end{tabular}} & No & 0.751 & 0.358 & 0.355 & 0.027 & 8.8 & 22.4 \\
& Yes & \textbf{0.939} & \textbf{0.742} & \textbf{0.545} & \textbf{0.375} & 31.8 & 94.2 \\
\multirow{2}{*}{\bf \begin{tabular}[c]{@{}c@{}}\pipeline \\ noDAG\end{tabular}} & No & 0.807 & 0.391 & 0.347 & 0.049 & 12.3 & 25.8 \\
& Yes & 0.936 & 0.681 & 0.417 & 0.353 & 32.0 & 98.5 \\
\multirow{2}{*}{\bf \fullproof} & No & / & / & 0.021 & 0.027 & 0.8 & 10.5 \\
& Yes & / & / & 0.123 & 0.141 & 10.7 & 76.7 \\
\multirow{2}{*}{\bf \steplevel} & No & / & 0.068 & 0.046 & 0.005 & 0.2 & 1.2 \\
& Yes & / & 0.105 & 0.072 & 0.005 & 10.5 & 71.6 \\
\end{tabular}
\end{center}
\vspace{-1.35\baselineskip}               
\end{table}



%% file: tab_error_analysis.tex

\begin{table}[t]
\vspace{-0.2cm}
\caption{Breakdown of step-level outcomes for different pipeline configurations. ``None" indicates the percentage of total steps completed successfully, while other columns show the percentage of steps that failed due to a specific error source.\vspace{-0.3cm}}
\label{tab:error_analysis_main}
\begin{center}
\begin{tabular}{ccccccc}
\multicolumn{1}{c}{\bf Pipeline} & \multicolumn{1}{c}{\bf Think} & \multicolumn{1}{c}{\bf Total Steps} & \multicolumn{4}{c}{\bf Error Source (\%)} \\
\cline{4-7}
& & & \multicolumn{1}{c}{\bf None} & \multicolumn{1}{c}{\bf Formalizer} & \multicolumn{1}{c}{\bf Tactic} & \multicolumn{1}{c}{\bf NL Statement} \\
\hline \\
\pipeline \ DAG & No & 1735 & 33.7 & 46.5 & 19.6 & 0.2 \\
\pipeline \ DAG & Yes & 1737 & \textbf{53.3} & 38.9 & 5.6 & 2.2 \\
\pipeline \ noDAG & No & 1751 & 32.8 & 45.6 & 21.4 & 0.2 \\
\pipeline \ noDAG & Yes & 1755 & 42.8 & 47.0 & 7.6 & 2.6 \\
\end{tabular}
\end{center}
\vspace{-1.50\baselineskip}               
\end{table}

%% file: 099appendix.tex
\subsection{LLM Usage Statement}

In this work, Large Language Models (LLMs) were employed in specific components of our research pipeline and experimental framework. The primary applications of LLMs include:

1. \textbf{\pipeline \ Pipeline Implementation}: LLMs were utilized as core components within our proposed ProofFlow pipeline as Graph Builder, Formalizer and Tactic Completer.

2. \textbf{Existing Methods Comparison}: During the experimental evaluation, LLMs were employed to perform inference with existing methods for comparative analysis against our proposed approach.

3. \textbf{\proofscore \ Computation}: LLMs were integrated into the calculation process of our proposed \proofscore \ metric for quantitative assessment of formalization quality.

4. \textbf{Benchmark Dataset Construction}: The proof graphs included in our benchmark dataset were initially generated by LLMs, followed by human verification.

LLMs served as computational tools in these specific applications and for grammatical correction of an early draft. All research ideation, methodological design, analysis, and interpretation of results were conducted exclusively by the human authors, who take full responsibility for the content of this paper, including any LLM-assisted components.

\subsection{Additional Details on \proofscore}
\label{sect:leanscore}

\subsubsection{Evaluation of semantic equivalence of \proofscore}
\label{app: leanscore_ours}
In this section, we detail how to obtain the semantic equivalence score $f_i$ of each node, as proposed in Section~\ref{sect:score}. Note that by default, the score $f_i$ is 0 if the syntactic check at the formalizer step is not passed.

In particular, we assess the semantic equivalence between the self-contained natural language statement $nls_i$ and Lean code $lc_i$ of each node, which are the input and output of the Formalizer step of \pipeline \ (Sect.~\ref{sect:pipeline}). The process is as follows:
\begin{enumerate}[leftmargin=*, itemsep=0.25em]
\item First, we prompt LLM to break down both $nls_i$ and $lc_i$ into components. 
\item We employ an LLM-as-a-judge to evaluate the semantic equivalence between each formalized component, $lc_i$, and its natural language counterpart, $nls_i$. The LLM is provided with few-shot examples illustrating specific error types and assigns one of the following evaluations to each component: \begin{itemize} \item \textit{Perfectly match}: The component $lc_i$ fully captures the reasoning of $nls_i$. It may include additional constraints or conditions that were implicit in the natural language but are necessary for formal rigor. \item \textit{Minor inconsistency}: The component $lc_i$ correctly represents the core logic of $nls_i$ but may feature slight structural reordering or other small deviations. \item \textit{Major inconsistency}: The component $lc_i$ either omits a key part of the logic from $nls_i$ or introduces entirely different reasoning. \end{itemize} \item A Fuzzy measurement score is then computed for each component based on these evaluations, following the method in~\cite{xuejun2025mathesis}. This score is effectively a weighted average of the counts for ``Perfectly match'' and ``Minor inconsistency,'' with zero tolerance for any component rated as a ``Major inconsistency.'' \item The final score, $f_i$, is calculated by aggregating the Fuzzy measurements using a Sugeno Integral~\citep{sugeno1974theory}. 
\end{enumerate}

In brief, the score $f_i$ of a proof step is equal to $1$ if the Lean code fully captures the self-contained natural language statement with all necessary conditions and $f_i$ is equal to $0$ if at least one component of the Lean code contains major flaws.

\subsubsection{\proofscore \  for \fullproof \  and \steplevel}
\label{app:leanscore_competitors}
Our evaluation score, \proofscore, is the average of faithfulness evaluations of individual steps (nodes in the dependency graph) of our \pipeline \ method. To create a fair comparison with \fullproof \ and \steplevel, we adapted this evaluation metric as follows. In all cases, any syntactically incorrect Lean code was automatically assigned a score of 0. 
\begin{itemize}[leftmargin=*, itemsep=0.25em] 
\item For \fullproof: Since this method generates a single, complete proof for each problem, we applied the scoring process (detailed in Appendix~\ref{app: leanscore_ours}) to the entire proof. The final score in Tables~\ref{table:comparison_table_1} and~\ref{table:comparison_table_2} is the average score across all problems. Note that if the proof fails Lean syntactic check, the score is $0$ by default.
\item For \steplevel: This method generates code for each individual step. Therefore, we applied the scoring process to each step independently. The final score in the tables is the average across all individual steps from every problem in the benchmark. If any step fails Lean syntactic check, the score is $0$ by default.
\end{itemize} 
The results of this comparative analysis are presented in Tables~\ref{table:comparison_table_1} and~\ref{table:comparison_table_2}.

\subsection{Benchmark dataset statistics}\label{sect:benchmark_stats}

\begin{figure}[H]
    \centering
    \includegraphics[width=1\linewidth]{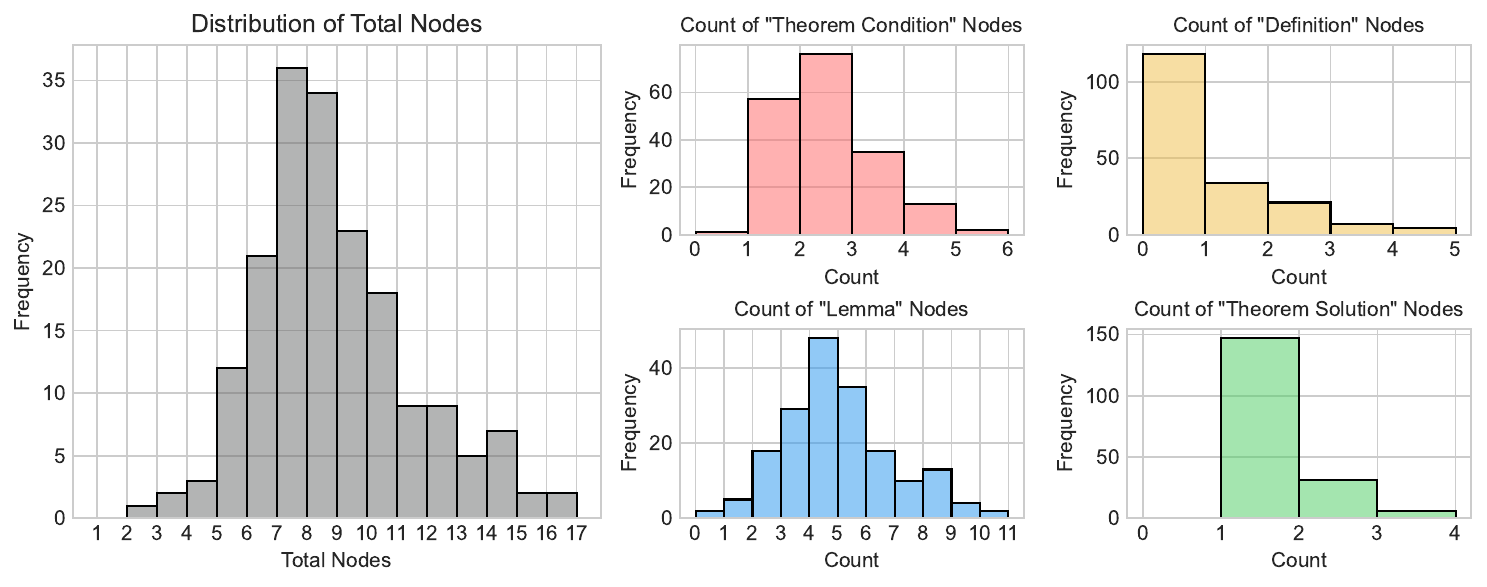}
    \caption{The distribution of the total number of nodes per problem (left) and the frequency of each node type (right).}
    \label{fig:benchmark_distribution}
\end{figure}

The distribution of the number of nodes per proof in the \benchmark \ benchmark is shown in Figure \ref{fig:benchmark_distribution}. On average, the proofs have 2 theorem conditions, 0.6 definitions, 4.4 lemmas, and 1.2 theorem solutions.

\subsection{Comparison results} \label{app:comparison_results}

This section contains extra comparison results. We show in Table~\ref{table:comparison_table_2} the extended version of Table~\ref{table:comparison_table_1}, which includes Pass@1, 3, and 5 rates.

\input{tab_comparison}

\subsection{Comparison Examples}  \label{app:examples}

In this section, we present several examples of proof autoformalization to illustrate how different pipelines perform when formalizing natural language proofs.  The examples below, selected from our comparative study in Section \ref{sect:comparison}, show how \pipeline \ DAG maintains a high degree of fidelity to the input proof. In contrast, other approaches, such as \pipeline \ noDAG, \fullproof, and \steplevel, often fail to either adhere to the flow of the natural proof or even generate valid Lean 4 code.

To evaluate structural fidelity, we analyzed the dependencies for each syntactically correct proof step. By inspecting the Lean tactics, we identified exactly which previously proven steps and theorem conditions the solver utilized. A step was deemed structurally faithful if this set of dependencies precisely matched the logic of the original natural language proof. If the dependencies differed in any way, the step was marked as unfaithful.

For the purposes of illustration, the figures presented in this section are based on the actual dependency graph of the input natural language proof, and annotations will be provided in the case the the logical flow of the generated proof graph deviates from this input dependency graph.

\subsubsection{Example: \pipeline \ DAG is superior in structure fidelity}

This example corresponds to entry ``dummy\_6" in the \benchmark \ benchmark.

\begin{figure}[t]
    \centering
    \includegraphics[width=1\linewidth]{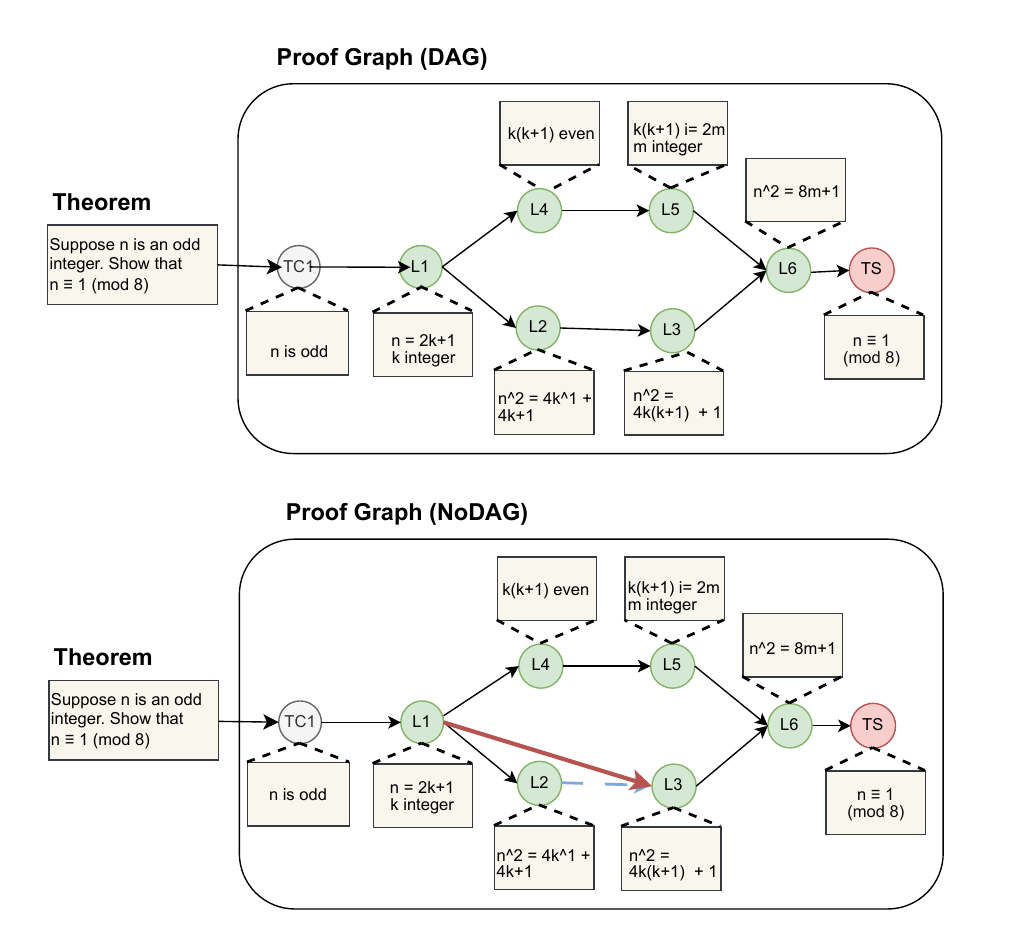}
    \caption{Comparison of proof structures generated by \pipeline \ DAG (top) and \pipeline \ noDAG (bottom) with respect to the original natural language proof. The red arrow indicates a dependency in \pipeline \ noDAG not in the natural language proof. The blue dashed arrow indicates a dependence in the natural language proof but not included in \pipeline \ noDAG. The \pipeline \ DAG  formal proof faithfully follows the dependency of the original proof. However, the one by \pipeline \ noDAG proves step \textsc{L2} then reuses \textsc{L1} to prove \textsc{L3}, which is structurally unfaithful to the original proof.}
    \label{fig:cheating_dummy6}
\end{figure}
\vspace{-2mm}

\begin{theorembox}
If $n$ is an odd integer, then $n^2 \equiv 1 \pmod{8}$.
\end{theorembox}

\begin{proofbox}
Since $n$ is odd, we can write $n = 2k + 1$ for some integer $k$. Then $n^2 = (2k + 1)^2 = 4k^2 + 4k + 1$. We can factor this as $n^2 = 4k(k + 1) + 1$. Now, either $k$ is even or $k$ is odd. If $k$ is even, then $k + 1$ is odd, and if $k$ is odd, then $k + 1$ is even. In either case, $k(k + 1)$ is even, so $k(k + 1) = 2m$ for some integer $m$. Therefore $n^2 = 4(2m) + 1 = 8m + 1$, which means $n^2 \equiv 1 \pmod{8}$.
\end{proofbox}

\textbf{\pipeline \ DAG:}
The proof produced by \pipeline \ DAG follows correctly the structure of the natural language proof and generates syntatically correct Lean code (see Figure~\ref{fig:cheating_dummy6}). 

\textbf{\pipeline \ noDAG:}
 \pipeline \ noDAG fails to adhere to the dependency structure of the natural language proof. In particular, as illustrated in Fig.~\ref{fig:cheating_dummy6}, the transition from step L2 to L3 was inherent in the natural language proof but neglected in noDAG. The proof code for step L3 utilized step L1 to reprove the result of step L2. In other words, step L2 was made redundant. 

\textbf{\fullproof:}
\fullproof \ failed to generate syntactically correct Lean code (the first syntax error is ``unknown identifier").

\textbf{\steplevel:}
\steplevel \ failed to generate syntactically correct Lean code for steps \textsc{L4, L5, L6}, and \textsc{TS} (the first syntax error was ``type mismatch").

\subsubsection{Example: \pipeline DAG is superior in prover accuracy}
This example corresponds to entry ``dummy\_7" in the \benchmark \ benchmark.

\begin{theorembox}
If $P(A) = 0.6$ and $P(B) = 0.7$, then $P(A \cap B) \geq 0.3$.
\end{theorembox}

\begin{proofbox}
We know that $P(A \cup B) = P(A) + P(B) - P(A \cap B)$. Since $P(A \cup B) \leq 1$, we have $P(A) + P(B) - P(A \cap B) \leq 1$. Substituting the given values: $0.6 + 0.7 - P(A \cap B) \leq 1$, which gives $1.3 - P(A \cap B) \leq 1$. Therefore $P(A \cap B) \geq 0.3$.
\end{proofbox}

\begin{figure}[h]
    \centering
    \includegraphics[width=1\linewidth]{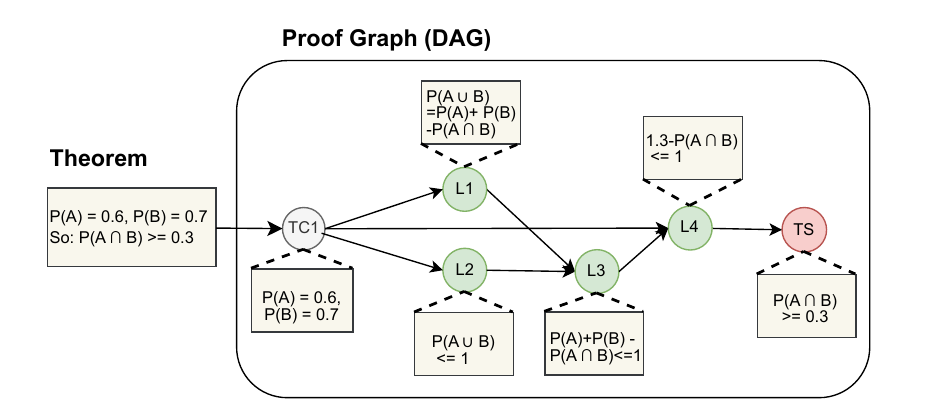}
    \caption{The structure of the proof produced by \pipeline \ DAG faithfully follows the dependency graph of the input natural language proof.}
    \label{fig:cheating_dummy7}
\end{figure}

\textbf{\pipeline \ DAG:}
\pipeline \ DAG generates step-by-step proof faithfully following the logical transition of the natural language proof (see Fig.~\ref{fig:cheating_dummy7}). 

\textbf{\pipeline \ noDAG:}
The Lean code proof achieves very low prover accuracy. This is likely attributed to the fact that each step is given all previous steps, in contrast to only the necessary steps in DAG, leading to confusion that the prover is obliged to utilize all previous steps. As a consequence, the prover fails to produce concise and correct Lean code.

\textbf{\fullproof:}
\fullproof \ failed to generate syntactically correct Lean code to prove this theorem (the first syntax error is ``unknown identifier").

\textbf{\steplevel:}
\steplevel \ failed to generate syntactically correct Lean code to prove this theorem (the first syntax error is function type error).

\subsubsection{Example: \pipeline \ DAG is superior in proof efficiency}
This example corresponds to entry ``dummy\_9" in the \benchmark \ benchmark.

\begin{theorembox}
If $(a_n)$ is an arithmetic sequence with $a_1 = 5$ and $a_3 = 11$, then $a_5 = 17$.
\end{theorembox}

\begin{proofbox}
Since $(a_n)$ is arithmetic, there exists a common difference $d$ such that $a_n = a_1 + (n-1)d$ for all $n$. From the given information, $a_3 = a_1 + 2d$. Substituting the values: $11 = 5 + 2d$, which gives us $2d = 6$, so $d = 3$. Now we can find $a_5 = a_1 + 4d = 5 + 4(3) = 5 + 12 = 17$.
\end{proofbox}

\begin{figure}[h]
    \centering
    \includegraphics[width=1\linewidth]{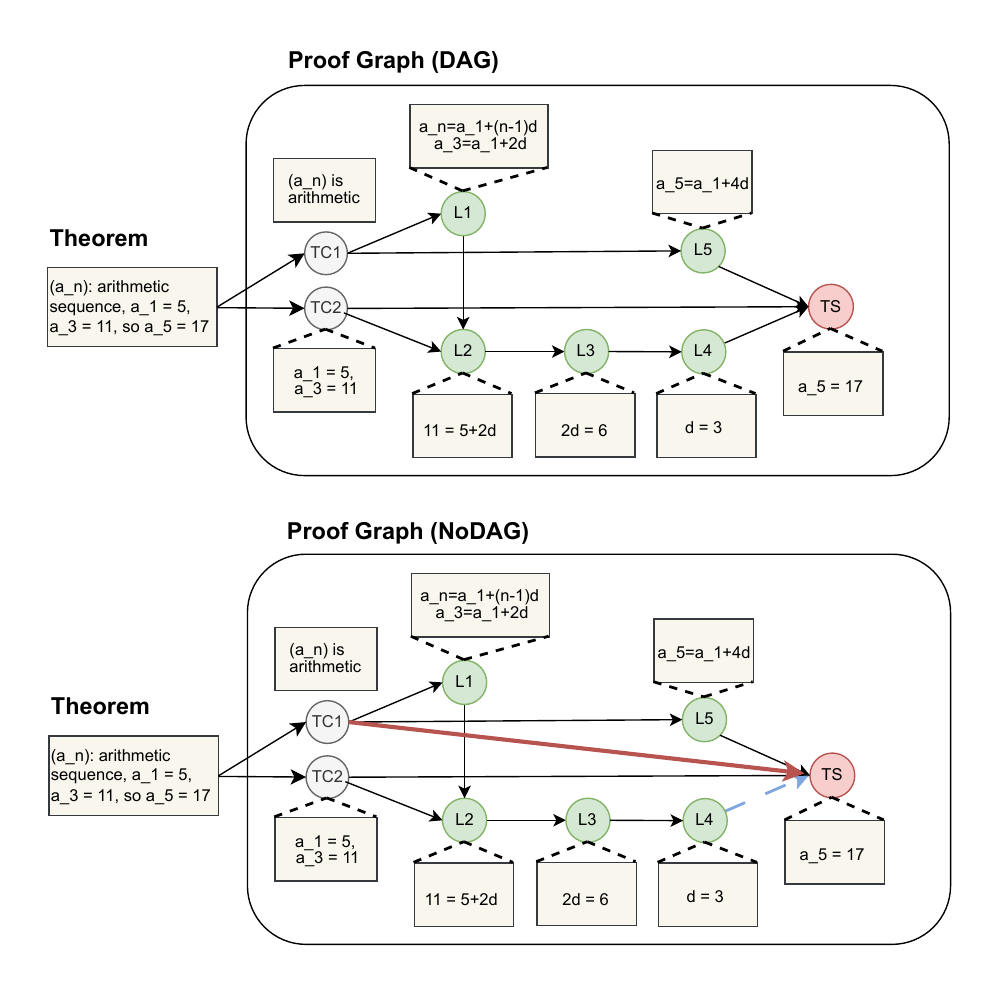}
    \caption{Comparison of proof structures generated by \pipeline \ DAG (top) and \pipeline \ noDAG (bottom) with respect to the original natural language proof. The red arrow indicates a dependency in \pipeline \ noDAG not in the natural language proof. The blue dashed arrow indicates a dependence in the natural language proof but servered in \pipeline \ noDAG. The one by \pipeline \ DAG faithfully follows the dependency of the original proof. However, the one by \pipeline \ noDAG proves the final step \textsc{TS} solely by the two given theorem conditions \textsc{TC1} and \textsc{TC2} and step \textsc{L5}. This is not only structurally unfaithful to the structure of the natural language proof but also inefficient as the efforts to prove all intermediate steps \textsc{L1, L2, L3, L4} are squandered.}
    \label{fig:cheating_dummy9}
\end{figure}

\textbf{\pipeline \ DAG:}
\pipeline \ DAG generates step-by-step proof faithfully following the logical transition of the natural language proof (Fig.~\ref{fig:cheating_dummy9}). 

\textbf{\pipeline \ noDAG:}
The Lean code proof by \pipeline \ noDAG is not only structurally unfaithful but also inefficient. As illustrated in Fig.~\ref{fig:cheating_dummy9}, the final step \textsc{TS} is proven by the tactics stemming from theorem conditions \textsc{TC1} and \textsc{TC2} and intermediate step \textsc{L5} (direct result of \textsc{TC1}) while neglecting all intermediate steps \textsc{L1, L2, L3, L4}. These steps are proven but not utilized for the final goal of the proof. As a consequence, the logical flow of the proof by \pipeline \ noDAG fails to adhere to the structural of the natural language proof and squanders the resources used to prove \textsc{L1, L2, L3, L4}.

\textbf{\fullproof:}
\fullproof \ failed to generate syntactically correct Lean code to prove this theorem (the first syntax error is function type error).

\textbf{\steplevel:}
\steplevel \ failed to generate syntactically correct Lean code to prove this theorem (the first syntax error is function type error).

\subsection{Examples of errors in the natural language proof} \label{sect:nl_errors}

The errors listed in Table~\ref{tab:error_analysis} were identified using the thinking mode of our \pipeline \ DAG pipeline at pass@5. These examples were specifically extracted from proof steps that our error detection pipeline (detailed in Section~\ref{sect:error_analysis_main}) flagged as an ``NL statement error."

We have decided to keep these issues in the benchmark because they reflect the kinds of common and subtle ambiguities made by humans when writing proofs. For example, using ambiguous informal language like ``we get [result x]" is a frequent occurrence in practice. By including these ambiguities, we ensure the benchmark remains a realistic representation of human-written proofs, rather than an overly artificial one.

\begin{table}[ht]
\renewcommand{\arraystretch}{1.25} 
\centering
\begin{tabular}{@{}p{0.4\textwidth}p{0.2\textwidth}p{0.35\textwidth}@{}}
Original Natural Language & Error Type & Comment \\
\hline
$3a^3 - 3a^2b - 3ab^2 + 3b^3 = (a^2 - b^2)(a-b)$ & Incorrect Statement & An outright algebraic error. The left-hand side is exactly three times the right-hand side. \\
Let~$n$ be a natural number. If $n=1$, ... If $n$ is prime, we are done. If $n$ is composite, then $n=ab$ ... By induction... & Logical Flaw & The proof sketch is intuitively correct but structurally flawed. It conflates the base case and the inductive step of a strong induction proof. \\
...we get $(r\frac{\sqrt{3}}{2})^2 + (r/2-1)^2 = 1$ & Ambiguity & The phrase ``we get'' misleadingly suggests a general identity, when this is actually a conditional equation that only holds for the specific value $r=1$. \\
From the condition $\operatorname{Arg}(z) = \pi/6$, we can write $z = r(\frac{\sqrt{3}}{2} + \frac{i}{2})$ for some $r>0$. & Missing Assumption & The statement is incomplete because it relies on the unstated precondition that $z \neq 0$ for the $\operatorname{Arg}(z)$ function to be well-defined. \\
Hence, $(X\cdot Y)^2\le|X|^2|Y|^2$, because if not, then p would have two distinct real zeros... & Incomplete Argument & The reasoning, based on the discriminant of a quadratic, is incomplete as it fails for the case where $Y=0$, where the polynomial degenerates. \\ 
Integrating... yields $v(x,y) = 2xy + 3y + g(x)$, where $g(x)$ is a function of $x$. & Incomplete Statement & The statement is true but insufficient. In the context of solving a PDE, it omits the crucial condition that the function of integration $g(x)$ must also be differentiable. \\
\hline
\end{tabular}
\caption{Analysis of Flaws Identified in Natural Language Proof Steps}
\label{tab:error_analysis}
\end{table}

%% file: tab_comparison.tex
\begin{table}[t]
\caption{Performance metrics for all pipelines, evaluated at Pass@1, 3, and 5 rates on our 184-problem benchmark. Entries marked with ``/" indicate not applicable.}
\label{table:comparison_table_2}
\begin{center}
\renewcommand{\arraystretch}{1.25}
\begin{tabular}{ccc|cc|lccc}
\multicolumn{1}{c}{} & \multicolumn{1}{c}{} & \multicolumn{1}{c|}{} & \multicolumn{2}{c|}{\bf Step-Level} & \multicolumn{4}{c}{\bf Proof-Level} \\
\multicolumn{1}{c}{\bf Pipeline} & \multicolumn{1}{c}{\bf \begin{tabular}[c]{@{}c@{}}Think \\ mode\end{tabular}} & \multicolumn{1}{c|}{\bf Pass} & \multicolumn{1}{c}{\bf \begin{tabular}[c]{@{}c@{}}Form. \\ accuracy\end{tabular}} & \multicolumn{1}{c|}{\bf \begin{tabular}[c]{@{}c@{}}Tactic \\ accuracy\end{tabular}} & \multicolumn{1}{c}{\bf \begin{tabular}[c]{@{}c@{}}Proof \\ Score\end{tabular}} & \multicolumn{1}{c}{\bf \begin{tabular}[c]{@{}c@{}}Correct \\ syntax\end{tabular}} & \multicolumn{1}{c}{\bf \begin{tabular}[c]{@{}c@{}}Time \\ (mins)\end{tabular}} & \multicolumn{1}{c}{\bf \begin{tabular}[c]{@{}c@{}}Output \\ tokens (k)\end{tabular}} \\ 
\hline \\
\multirow{6}{*}{\begin{tabular}[c]{@{}c@{}}\pipeline\\ DAG\end{tabular}} & \multirow{3}{*}{No} & 1 & 0.644 & 0.252 & 0.320 & 0.016 & 3.3 & 8.1 \\
 & & 3 & 0.722 & 0.323 & 0.347 & 0.027 & 6.3 & 15.7 \\
 & & 5 & 0.751 & 0.358 & 0.355 & 0.027 & 8.8 & 22.4 \\
 & \multirow{3}{*}{Yes} & 1 & 0.844 & 0.629 & 0.508 & 0.245 & 19.3 & 55.7 \\
 & & 3 & 0.925 & 0.723 & 0.541 & 0.348 & 27.0 & 78.0 \\
 & & 5 & \textbf{0.939} & \textbf{0.742} & \textbf{0.545} & \textbf{0.375} & 31.8 & 94.2 \\
 \\
\multirow{6}{*}{\begin{tabular}[c]{@{}c@{}}\pipeline\\ noDAG\end{tabular}} & \multirow{3}{*}{No} & 1 & 0.697 & 0.225 & 0.312 & 0.022 & 4.8 & 9.8 \\
 & & 3 & 0.791 & 0.328 & 0.344 & 0.038 & 9.0 & 18.4 \\
 & & 5 & 0.807 & 0.391 & 0.347 & 0.049 & 12.3 & 25.8 \\
 & \multirow{3}{*}{Yes} & 1 & 0.860 & 0.573 & 0.397 & 0.217 & 18.9 & 56.0 \\
 & & 3 & 0.921 & 0.662 & 0.414 & 0.332 & 27.0 & 80.0 \\
 & & 5 & 0.936 & 0.681 & 0.417 & 0.353 & 32.0 & 98.5 \\
 \\
\multirow{6}{*}{\centering \fullproof} & \multirow{3}{*}{No} & 1 & / & / & 0 & 0 & 0.2 & 2.7 \\
 & & 3 & / & / & 0.011 & 0.011 & 0.5 & 6.6 \\
 & & 5 & / & / & 0.021 & 0.027 & 0.8 & 10.5 \\
 & \multirow{3}{*}{Yes} & 1 & / & / & 0.009 & 0.011 & 2.3 & 16.0 \\
 & & 3 & / & / & 0.065 & 0.082 & 6.7 & 48.2 \\
 & & 5 & / & / & 0.123 & 0.141 & 10.7 & 76.7 \\
 \\
\multirow{6}{*}{\steplevel} & \multirow{3}{*}{No} & 1 & / & 0.054 & 0 & 0 & 0.03 & 0.1 \\
 & & 3 & / & 0.061 & 0 & 0 & 0.05 & 0.3 \\
 & & 5 & / & 0.068 & 0.046 & 0.005 & 0.2 & 1.2 \\
 & \multirow{3}{*}{Yes} & 1 & / & 0.068 & 0 & 0 & 0.9 & 5.9 \\
 & & 3 & / & 0.093 & 0 & 0 & 1.7 & 11.6 \\
 & & 5 & / & 0.105 & 0.072 & 0.005 & 10.5 & 71.6 \\
\end{tabular}
\end{center}
\end{table}

%% file: arXiv_version.bbl
\begin{thebibliography}{40}
\providecommand{\natexlab}[1]{#1}
\providecommand{\url}[1]{\texttt{#1}}
\expandafter\ifx\csname urlstyle\endcsname\relax
  \providecommand{\doi}[1]{doi: #1}\else
  \providecommand{\doi}{doi: \begingroup \urlstyle{rm}\Url}\fi

\bibitem[Baba et~al.(2025)Baba, Liu, Kurita, and Sannai]{baba2025prover}
Kaito Baba, Chaoran Liu, Shuhei Kurita, and Akiyoshi Sannai.
\newblock Prover agent: An agent-based framework for formal mathematical proofs.
\newblock \emph{arXiv preprint arXiv:2506.19923}, 2025.

\bibitem[Barras et~al.(1997)Barras, Boutin, Cornes, Courant, Filli{\^{a}}tre, Gim{\'{e}}nez, Herbelin, Mohring, Sa{\"{i}}bi, and Werner]{barras1997coq}
Bruno Barras, Samuel Boutin, Cristina Cornes, Judica{\"{e}}l Courant, Jean-Christophe Filli{\^{a}}tre, Eduardo Gim{\'{e}}nez, H{\'{e}}ctor Herbelin, G{\'{e}}rard Mohring, Amokrane Sa{\"{i}}bi, and Benjamin Werner.
\newblock {The Coq Proof Assistant Reference Manual: Version 6.1}.
\newblock Technical Report RT-0203, INRIA, 1997.

\bibitem[Buzzard \& Taylor(2025)Buzzard and Taylor]{buzzard2025lean}
Kevin Buzzard and Richard Taylor.
\newblock {A Lean proof of Fermat’s Last Theorem}.
\newblock Technical report, Technical report, Imperial College of London, 2024. https~…, 2025.

\bibitem[Cao et~al.(2025)Cao, Song, Li, Le, Zhang, Xue, and Yang]{cao2025reviving}
Chenrui Cao, Liangcheng Song, Zenan Li, Xinyi Le, Xian Zhang, Hui Xue, and Fan Yang.
\newblock {Reviving DSP for Advanced Theorem Proving in the Era of Reasoning Models}.
\newblock \emph{arXiv preprint arXiv:2506.11487}, 2025.

\bibitem[Chen et~al.(2025)Chen, Gu, Huang, Huang, Jiang, Jie, Jin, Jin, Li, Ma, et~al.]{chen2025seed}
Luoxin Chen, Jinming Gu, Liankai Huang, Wenhao Huang, Zhicheng Jiang, Allan Jie, Xiaoran Jin, Xing Jin, Chenggang Li, Kaijing Ma, et~al.
\newblock Seed-prover: Deep and broad reasoning for automated theorem proving.
\newblock \emph{arXiv preprint arXiv:2507.23726}, 2025.

\bibitem[Cunningham et~al.(2023)Cunningham, Bunescu, and Juedes]{cunningham2023towards}
Garett Cunningham, Razvan~C Bunescu, and David Juedes.
\newblock Towards autoformalization of mathematics and code correctness: Experiments with elementary proofs.
\newblock \emph{arXiv preprint arXiv:2301.02195}, 2023.

\bibitem[Gao et~al.(2024)Gao, Wang, Jiang, Gao, Qin, Xu, and Dong]{gao2024herald}
Guoxiong Gao, Yutong Wang, Jiedong Jiang, Qi~Gao, Zihan Qin, Tianyi Xu, and Bin Dong.
\newblock {Herald: A natural language annotated lean 4 dataset}.
\newblock \emph{arXiv preprint arXiv:2410.10878}, 2024.

\bibitem[Gonthier et~al.(2013)Gonthier, Asperti, Avigad, Bertot, Cohen, Garillot, Le~Roux, Mahboubi, O'Connor, Biha, et~al.]{gonthier2013machine}
Georges Gonthier, Andrea Asperti, Jeremy Avigad, Yves Bertot, Cyril Cohen, Fran{\c{c}}ois Garillot, St{\'e}phane Le~Roux, Assia Mahboubi, Russell O'Connor, Sidi~Ould Biha, et~al.
\newblock {A machine-checked proof of the Odd Order Theorem}.
\newblock \emph{Lecture Notes in Computer Science}, 8332:\penalty0 163--179, 2013.

\bibitem[Hales et~al.(2017)Hales, Adams, Bauer, Dang, Harrison, Hoang, Kaliszyk, Magron, McLaughlin, Nguyen, et~al.]{hales2017formal}
Thomas Hales, Mark Adams, Gertrud Bauer, Tat~Dat Dang, John Harrison, Le~Truong Hoang, Cezary Kaliszyk, Victor Magron, Sean McLaughlin, Tat~Thang Nguyen, et~al.
\newblock {A formal proof of the Kepler conjecture}.
\newblock \emph{Forum of Mathematics, Pi}, 5, 2017.

\bibitem[Hu et~al.(2025)Hu, Zhou, Grechuk, and Tyukin]{hu2025stepproofstepbystepverificationnatural}
Xiaolin Hu, Qinghua Zhou, Bogdan Grechuk, and Ivan~Y. Tyukin.
\newblock Stepproof: Step-by-step verification of natural language mathematical proofs, 2025.
\newblock URL \url{https://arxiv.org/abs/2506.10558}.

\bibitem[Huang et~al.(2025)Huang, Jin, Liang, Li, and Liu]{huang2025formarl}
Yanxing Huang, Xinling Jin, Sijie Liang, Peng Li, and Yang Liu.
\newblock Formarl: Enhancing autoformalization with no labeled data.
\newblock \emph{arXiv preprint arXiv:2508.18914}, 2025.

\bibitem[Jiang et~al.(2023)Jiang, Welleck, Zhou, Li, Liu, Jamnik, Lacroix, Wu, and Lample]{jiang2023draftsketchproveguiding}
Albert~Q. Jiang, Sean Welleck, Jin~Peng Zhou, Wenda Li, Jiacheng Liu, Mateja Jamnik, Timothée Lacroix, Yuhuai Wu, and Guillaume Lample.
\newblock Draft, sketch, and prove: Guiding formal theorem provers with informal proofs, 2023.
\newblock URL \url{https://arxiv.org/abs/2210.12283}.

\bibitem[Jiang et~al.(2024)Jiang, Li, Jamnik, Holden, and Paulson]{jiang2023draft}
Albert~Q. Jiang, Wenda Li, Mateja Jamnik, Sean~B. Holden, and Lawrence~C. Paulson.
\newblock Draft, sketch, and prove: Guiding formal theorem provers with informal proofs.
\newblock In \emph{The Twelfth International Conference on Learning Representations (ICLR)}, 2024.
\newblock URL \url{https://openreview.net/forum?id=Vp0Zz0w2V1}.

\bibitem[Liang et~al.(2025)Liang, Song, Li, Yang, Zhang, Mi, and Yu]{liang2025towards}
Zhenwen Liang, Linfeng Song, Yang Li, Tao Yang, Feng Zhang, Haitao Mi, and Dong Yu.
\newblock Towards solving more challenging imo problems via decoupled reasoning and proving.
\newblock \emph{arXiv preprint arXiv:2507.06804}, 2025.

\bibitem[Lin et~al.(2025)Lin, Tang, Lyu, Yang, Chung, Zhao, Jiang, Geng, Ge, Sun, Wu, Gesi, Lu, Acuna, Yang, Lin, Choi, Chen, Arora, and Jin]{lin2025goedelproverv2scalingformaltheorem}
Yong Lin, Shange Tang, Bohan Lyu, Ziran Yang, Jui-Hui Chung, Haoyu Zhao, Lai Jiang, Yihan Geng, Jiawei Ge, Jingruo Sun, Jiayun Wu, Jiri Gesi, Ximing Lu, David Acuna, Kaiyu Yang, Hongzhou Lin, Yejin Choi, Danqi Chen, Sanjeev Arora, and Chi Jin.
\newblock Goedel-prover-v2: Scaling formal theorem proving with scaffolded data synthesis and self-correction, 2025.
\newblock URL \url{https://arxiv.org/abs/2508.03613}.

\bibitem[Liu et~al.(2025)Liu, Bao, Zhang, Liu, Liu, Chen, Jiao, and Luo]{liu2025atlas}
Xiaoyang Liu, Kangjie Bao, Jiashuo Zhang, Yunqi Liu, Yuntian Liu, Yu~Chen, Yang Jiao, and Tao Luo.
\newblock Atlas: Autoformalizing theorems through lifting, augmentation, and synthesis of data.
\newblock \emph{arXiv preprint arXiv:2502.05567}, 2025.

\bibitem[Lu et~al.(2024)Lu, Wan, Liu, Huang, Xiong, Liu, Shen, Jin, Zhang, Wang, et~al.]{lu2024process}
Jianqiao Lu, Yingjia Wan, Zhengying Liu, Yinya Huang, Jing Xiong, Chengwu Liu, Jianhao Shen, Hui Jin, Jipeng Zhang, Haiming Wang, et~al.
\newblock {Process-driven autoformalization in lean 4}.
\newblock \emph{arXiv preprint arXiv:2406.01940}, 2024.

\bibitem[{Math Inc.}(2025)]{strongpnt_github}
{Math Inc.}
\newblock The strong prime number theorem.
\newblock \url{https://github.com/math-inc/strongpnt}, 2025.
\newblock Accessed: 2025-09-11.

\bibitem[Moura \& Ullrich(2021)Moura and Ullrich]{moura2021lean}
Leonardo~de Moura and Sebastian Ullrich.
\newblock {The Lean 4 theorem prover and programming language}.
\newblock In \emph{International Conference on Automated Deduction}, pp.\  625--635. Springer, 2021.

\bibitem[Patel et~al.(2023)Patel, Saha, and Flanigan]{patel2023new}
Nilay Patel, Rahul Saha, and Jeffrey Flanigan.
\newblock A new approach towards autoformalization.
\newblock \emph{arXiv preprint arXiv:2310.07957}, 2023.

\bibitem[Pathak(2024)]{pathak2024gflean}
Shashank Pathak.
\newblock {GFLean: An autoformalisation framework for lean via GF}.
\newblock \emph{arXiv preprint arXiv:2404.01234}, 2024.

\bibitem[Paulson(1994)]{Paulson1994Isabelle}
Lawrence~C. Paulson.
\newblock Isabelle: A generic theorem prover.
\newblock In Alan Bundy (ed.), \emph{Proceedings of the 12th International Conference on Automated Deduction (CADE-12)}, volume 814 of \emph{Lecture Notes in Computer Science}, pp.\  37--41. Springer, 1994.

\bibitem[Poiroux et~al.(2024)Poiroux, Weiss, Kun{\v{c}}ak, and Bosselut]{poiroux2024improving}
Auguste Poiroux, Gail Weiss, Viktor Kun{\v{c}}ak, and Antoine Bosselut.
\newblock Improving autoformalization using type checking.
\newblock \emph{arXiv preprint arXiv:2406.07222}, 2024.

\bibitem[Ren et~al.(2025{\natexlab{a}})Ren, Shao, Song, Xin, Wang, Zhao, Zhang, Fu, Zhu, Yang, Wu, Gou, Ma, Tang, Liu, Gao, Guo, and Ruan]{ren2025deepseekproverv2advancingformalmathematical}
Z.~Z. Ren, Zhihong Shao, Junxiao Song, Huajian Xin, Haocheng Wang, Wanjia Zhao, Liyue Zhang, Zhe Fu, Qihao Zhu, Dejian Yang, Z.~F. Wu, Zhibin Gou, Shirong Ma, Hongxuan Tang, Yuxuan Liu, Wenjun Gao, Daya Guo, and Chong Ruan.
\newblock Deepseek-prover-v2: Advancing formal mathematical reasoning via reinforcement learning for subgoal decomposition, 2025{\natexlab{a}}.
\newblock URL \url{https://arxiv.org/abs/2504.21801}.

\bibitem[Ren et~al.(2025{\natexlab{b}})Ren, Shao, Song, Xin, Wang, Zhao, Zhang, Fu, Zhu, Yang, et~al.]{ren2025deepseek}
ZZ~Ren, Zhihong Shao, Junxiao Song, Huajian Xin, Haocheng Wang, Wanjia Zhao, Liyue Zhang, Zhe Fu, Qihao Zhu, Dejian Yang, et~al.
\newblock Deepseek-prover-v2: Advancing formal mathematical reasoning via reinforcement learning for subgoal decomposition.
\newblock \emph{arXiv preprint arXiv:2504.21801}, 2025{\natexlab{b}}.

\bibitem[Santos et~al.(2025)Santos, Wang, de~Saxcé, Wang, Baksys, Unsal, Liu, Liu, and Li]{santos2025kiminaleanservertechnical}
Marco~Dos Santos, Haiming Wang, Hugues de~Saxcé, Ran Wang, Mantas Baksys, Mert Unsal, Junqi Liu, Zhengying Liu, and Jia Li.
\newblock {Kimina Lean Server: Technical Report}, 2025.
\newblock URL \url{https://arxiv.org/abs/2504.21230}.

\bibitem[Scholze(2021)]{scholze2021liquid}
Peter Scholze.
\newblock {The Liquid Tensor Experiment}.
\newblock Xena Blog, 2021.
\newblock Available at: \url{https://xenaproject.wordpress.com/2021/05/18/the-liquid-tensor-experiment/}.

\bibitem[Shang et~al.(2025)Shang, Wan, Peng, Wu, Chen, Yan, and Zhang]{shang2025stepfun}
Shijie Shang, Ruosi Wan, Yue Peng, Yutong Wu, Xiong-hui Chen, Jie Yan, and Xiangyu Zhang.
\newblock Stepfun-prover preview: Let's think and verify step by step.
\newblock \emph{arXiv preprint arXiv:2507.20199}, 2025.

\bibitem[Sheng et~al.(2025)Sheng, Lyu, Jin, Xia, Gu, Zou, and Lu]{sheng2025solving}
Jiayi Sheng, Luna Lyu, Jikai Jin, Tony Xia, Alex Gu, James Zou, and Pan Lu.
\newblock Solving inequality proofs with large language models.
\newblock \emph{arXiv preprint arXiv:2506.07927}, 2025.

\bibitem[Sugeno(1974)]{sugeno1974theory}
Michio Sugeno.
\newblock Theory of fuzzy integrals and its applications.
\newblock \emph{Doctoral Thesis, Tokyo Institute of Technology}, 1974.

\bibitem[Tao(2023)]{tao2023pfr}
Terence Tao.
\newblock {Formalizing the proof of PFR in Lean4 using blueprint: a short tour}.
\newblock What's new Blog, 2023.
\newblock Available at: \url{https://terrytao.wordpress.com/2023/11/18/formalizing-the-proof-of-pfr-in-lean4-using-blueprint-a-short-tour/}.

\bibitem[Wang et~al.(2025{\natexlab{a}})Wang, Unsal, Lin, Baksys, Liu, Santos, Sung, Vinyes, Ying, Zhu, Lu, Saxcé, Bailey, Song, Xiao, Zhang, Zhang, Pu, Zhu, Liu, Bayer, Michel, Yu, Dreyfus-Schmidt, Tunstall, Pagani, Machado, Bourigault, Wang, Polu, Barroyer, Li, Niu, Fleureau, Hu, Yu, Wang, Yang, Liu, and Li]{kimina_prover_2025}
Haiming Wang, Mert Unsal, Xiaohan Lin, Mantas Baksys, Junqi Liu, Marco~Dos Santos, Flood Sung, Marina Vinyes, Zhenzhe Ying, Zekai Zhu, Jianqiao Lu, Hugues~de Saxcé, Bolton Bailey, Chendong Song, Chenjun Xiao, Dehao Zhang, Ebony Zhang, Frederick Pu, Han Zhu, Jiawei Liu, Jonas Bayer, Julien Michel, Longhui Yu, Léo Dreyfus-Schmidt, Lewis Tunstall, Luigi Pagani, Moreira Machado, Pauline Bourigault, Ran Wang, Stanislas Polu, Thibaut Barroyer, Wen-Ding Li, Yazhe Niu, Yann Fleureau, Yangyang Hu, Zhouliang Yu, Zihan Wang, Zhilin Yang, Zhengying Liu, and Jia Li.
\newblock Kimina-prover preview: Towards large formal reasoning models with reinforcement learning.
\newblock 2025{\natexlab{a}}.
\newblock URL \url{http://arxiv.org/abs/2504.11354}.

\bibitem[Wang et~al.(2025{\natexlab{b}})Wang, Unsal, Lin, Baksys, Liu, Santos, Sung, Vinyes, Ying, Zhu, et~al.]{wang2025kimina}
Haiming Wang, Mert Unsal, Xiaohan Lin, Mantas Baksys, Junqi Liu, Marco~Dos Santos, Flood Sung, Marina Vinyes, Zhenzhe Ying, Zekai Zhu, et~al.
\newblock Kimina-prover preview: Towards large formal reasoning models with reinforcement learning.
\newblock \emph{arXiv preprint arXiv:2504.11354}, 2025{\natexlab{b}}.

\bibitem[Welleck et~al.(2021)Welleck, Liu, Bras, Hajishirzi, Choi, and Cho]{welleck2021naturalproofsmathematicaltheoremproving}
Sean Welleck, Jiacheng Liu, Ronan~Le Bras, Hannaneh Hajishirzi, Yejin Choi, and Kyunghyun Cho.
\newblock Naturalproofs: Mathematical theorem proving in natural language, 2021.
\newblock URL \url{https://arxiv.org/abs/2104.01112}.

\bibitem[Wu et~al.(2022)Wu, Jiang, Li, Rabe, Staats, Jamnik, and Szegedy]{wu2022autoformalization}
Yuhuai Wu, Albert~Qiaochu Jiang, Wenda Li, Markus Rabe, Charles Staats, Mateja Jamnik, and Christian Szegedy.
\newblock Autoformalization with large language models.
\newblock \emph{Advances in neural information processing systems}, 35:\penalty0 32353--32368, 2022.

\bibitem[Wu et~al.(2025)Wu, Huang, Wan, Peng, Shang, Cao, Qi, Zhang, Du, Yan, et~al.]{wu2025stepfun}
Yutong Wu, Di~Huang, Ruosi Wan, Yue Peng, Shijie Shang, Chenrui Cao, Lei Qi, Rui Zhang, Zidong Du, Jie Yan, et~al.
\newblock Stepfun-formalizer: Unlocking the autoformalization potential of llms through knowledge-reasoning fusion.
\newblock \emph{arXiv preprint arXiv:2508.04440}, 2025.

\bibitem[Yu et~al.(2025{\natexlab{a}})Yu, Zhong, Feng, Zhai, Yousefzadeh, Ng, Liu, Shou, Xiong, Zhou, et~al.]{xuejun2025mathesis}
Xuejun Yu, Jianyuan Zhong, Zijin Feng, Pengyi Zhai, Roozbeh Yousefzadeh, Wei~Chong Ng, Haoxiong Liu, Ziyi Shou, Jing Xiong, Yudong Zhou, et~al.
\newblock Mathesis: Towards formal theorem proving from natural languages.
\newblock \emph{arXiv preprint arXiv:2506.07047}, 2025{\natexlab{a}}.

\bibitem[Yu et~al.(2025{\natexlab{b}})Yu, Peng, Ding, Li, Peng, Liu, Zhang, Yuan, Xin, Huang, et~al.]{yu2025formalmath}
Zhouliang Yu, Ruotian Peng, Keyi Ding, Yizhe Li, Zhongyuan Peng, Minghao Liu, Yifan Zhang, Zheng Yuan, Huajian Xin, Wenhao Huang, et~al.
\newblock Formalmath: Benchmarking formal mathematical reasoning of large language models.
\newblock \emph{arXiv preprint arXiv:2505.02735}, 2025{\natexlab{b}}.

\bibitem[Zhou et~al.(2024)Zhou, Staats, Li, Szegedy, Weinberger, and Wu]{zhou2024donttrustverify}
Jin~Peng Zhou, Charles Staats, Wenda Li, Christian Szegedy, Kilian~Q. Weinberger, and Yuhuai Wu.
\newblock Don't trust: Verify -- grounding llm quantitative reasoning with autoformalization, 2024.
\newblock URL \url{https://arxiv.org/abs/2403.18120}.

\bibitem[Zhou et~al.(2025)Zhou, Zhao, Zhang, Wang, Wang, Chen, Wang, Chen, Jie, Zhang, et~al.]{zhou2025solving}
Yichi Zhou, Jianqiu Zhao, Yongxin Zhang, Bohan Wang, Siran Wang, Luoxin Chen, Jiahui Wang, Haowei Chen, Allan Jie, Xinbo Zhang, et~al.
\newblock Solving formal math problems by decomposition and iterative reflection.
\newblock \emph{arXiv preprint arXiv:2507.15225}, 2025.

\end{thebibliography}
